\pdfoutput=1

\documentclass[11pt]{article}

\usepackage{EMNLP2022}

\usepackage{times}
\usepackage{latexsym}

\usepackage[T1]{fontenc}

\usepackage[utf8]{inputenc}

\usepackage{microtype}

\usepackage{inconsolata}

\usepackage{textcomp}
\usepackage{stfloats}
\usepackage{url}
\usepackage{verbatim}
\usepackage{graphicx}
\usepackage{algorithm}
\usepackage{algorithmic}
\usepackage{amssymb}
\usepackage{helvet}  
\usepackage{courier}  
\usepackage{newfloat}
\usepackage{listings}
\usepackage{multirow}
\usepackage{amsmath}
\usepackage{booktabs}
\usepackage{pifont}
\usepackage{color}

\newcommand{{\model}}{SpanProto}

\title{\model: A Two-stage Span-based Prototypical Network for \\ Few-shot Named Entity Recognition}

\author{Jianing Wang$^{1}$\thanks{\ \ \ \ J. Wang and C. Han contributed equally to this work.}, Chengcheng Han$^{1}$\footnotemark[1], Chengyu Wang$^{2}$, Chuanqi Tan$^{2}$, Minghui Qiu$^{2}$, \\
\bf{Songfang Huang$^{2}$, Jun Huang$^{2}$, Ming Gao$^{1,3}$\thanks{\ \ \ Corresponding author.}} \\
$^{1}$ School of Data Science and Engineering, East China Normal University, Shanghai, China \\
$^{2}$ Alibaba Group, Hangzhou, China\\
$^{3}$ KLATASDS-MOE, School of Statistics, East China Normal University, Shanghai, China \\
\texttt{lygwjn@gmail.com, 52215903007@stu.ecnu.edu.cn}, \\
\texttt{\{chengyu.wcy,chuanqi.tcq,minghui.qmh\}@alibaba-inc.com}, \\
\texttt{\{songfang.hsf,huangjun.hj\}@alibaba-inc.com}, \\
\texttt{mgao@dase.ecnu.edu.cn}
}
\begin{document}
\maketitle
\begin{abstract}
Few-shot Named Entity Recognition (NER) aims to identify named entities with very little annotated data. Previous methods solve this problem 
based on token-wise classification, which ignores the information of entity boundaries, and inevitably the performance is affected by the massive non-entity tokens.
To this end, we propose a seminal span-based prototypical network ({\model}) that tackles few-shot NER via a~\emph{two-stage} approach, including span extraction and mention classification. 
In the span extraction stage, we transform the sequential tags into a global boundary matrix, enabling the model to focus on the explicit boundary information.
For mention classification, we leverage prototypical learning to capture the semantic representations for each labeled span and make the model better adapt to novel-class entities. To further improve the model performance, we split out the false positives generated by the span extractor but not labeled in the current episode set, and then present a margin-based loss to separate them from each prototype region.
Experiments over multiple benchmarks demonstrate that our model outperforms strong baselines by a large margin.
\footnote{All the codes and datasets will be released to the EasyNLP framework~\cite{DBLP:journals/corr/abs-2205-00258}. URL:~\url{https://github.com/alibaba/EasyNLP}}
\end{abstract}

\section{Introduction}

\begin{figure}
\centering
\includegraphics[width=\linewidth]{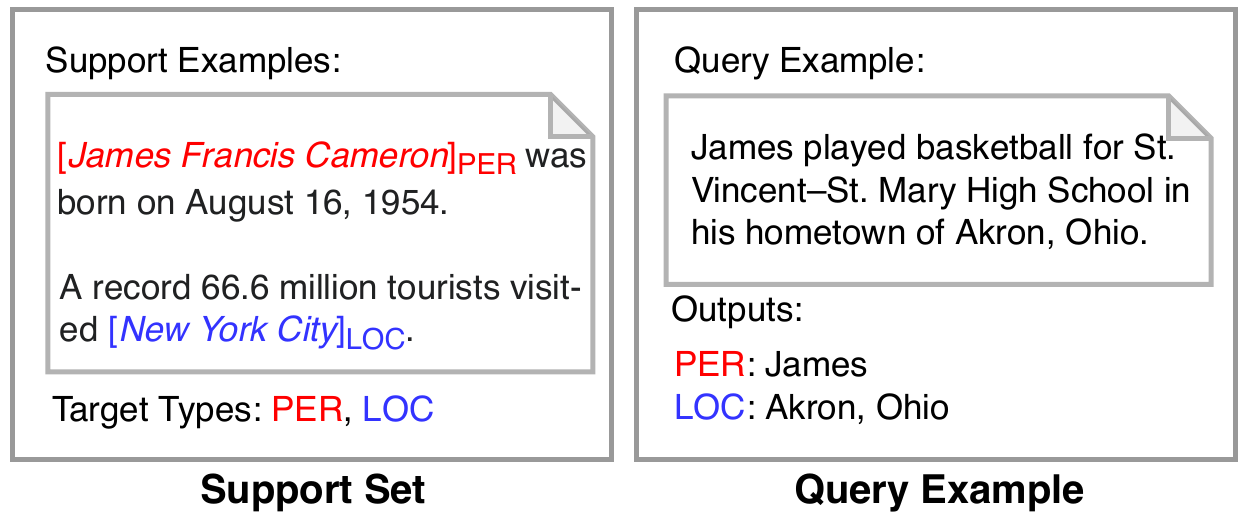}
\caption{An example of the 2-way 1-shot NER problem. Given a support set with 2 target types where each type is associated with 1 labeled entity, the task is to identify entities in the query example.}
\label{fig:example}
\end{figure}

Named Entity Recognition (NER) is one of the crucial tasks in natural language processing (NLP), which aims at extracting mention spans and classifying them into a set of pre-defined entity type classes. Previous methods present multiple deep neural architectures and achieve impressive performance~\cite{Huang2015Bidirectional, Santoro2016Meta, Ma2016End, Lample2016Neural, Matthew2017Semi}.
Yet, these conventional approaches heavily depend on the time-consuming and labor-intensive process of data annotation. 
Thus, an attractive problem of few-shot NER~\cite{Ding2020Few, Huang2021Few, Ma2022Decomposed} has been introduced which involves recognizing novel-class entities based on very few labeled examples (i.e. support examples).
An example of the 2-way 1-shot NER problem is shown in Figure~\ref{fig:example}.

To solve the
problem, multiple methods~\cite{Das2022CONTaiNER, Hou2020Few, Ziyadi2020Example, Fritzler2019Few, Ma2022Label}
follow the sequence labeling strategy that directly categorizes the entity type at the token level by calculating the distance between each query token and the prototype of each entity type class or the support tokens. However, the effect of these approaches are disturbed by numerous non-entity tokens (i.e. ``O'' class) and the token-wise label dependency (i.e. ``BIO'' rules)~\cite{Wang2021An,Shen2021Locate}. 
To bypass these issues, a branch of~\emph{two-stage} methods arise to decompose NER into two separate processes~\cite{Shen2021Locate, Wang2021An, Ma2022Decomposed, Wu2022Propose}, including span extraction and mention classification. Specifically, they first extract multiple spans via a class-agnostic model and then assign the label for each predicted span based on metric learning. 
Despite the success, there could still be two remaining problems. 1) The performance of span extraction is the upper limit for the whole system, which is still far away from satisfaction. 2) Previous methods ignore false positives generated during span extraction. 
Intuitively, because the decomposed model is class-agnostic in the span extraction stage, it could generate some entities which have no available entity type to be assigned in the target type set. In Figure~\ref{fig:example}, the model could extract a span of ``August 16, 1954'' (may be an entity of time type in another episode), yet, the existing methods still assign it a label as ``PER'' or ``LOC''~\footnote{Previous works suppose that all spans generated by the span extractor can be assigned with a type, which is unrealistic in the two-stage scenario.}.

To address these limitations, we present a novel~\textbf{Span}-based~\textbf{Proto}typical Network ({\model}) via a~\emph{two-stage} approach.
For span extraction, we introduce a~\emph{Span Extractor}, which aims to find the candidate spans. Different from the recent work~\cite{Ma2022Decomposed} which models it as a sequence labeling task, we convert the sequential tags to a global boundary matrix, which denotes the sentence-level target label, enabling the model to learn the explicit span boundary information regardless of the token-wise label dependency. 
For mention classification, we propose a~\emph{Mention Classifier} which aims at assigning a pre-defined entity type for each recalled span. 
When training the mention classifier, we compute the prototype embeddings for each entity type class based on the support examples, and leverage the prototypical learning to adjust the span representations in the semantic space.
To address the problem of false positives, we additionally design a margin-based loss to enlarge the semantic distance between false positives and all prototypes.
We conduct extensive experiments over multiple benchmarks, including Few-NERD~\cite{Ding2020Few} and CrossNER~\cite{Hou2020Few}. Results show that our method consistently outperforms state-of-the-art baselines 
by a large margin.
We summarize our main contributions as follows:
\begin{itemize}
    \item We propose a novel~\emph{two-stage} framework named {SpanProto} that solves the problem of few-shot NER with two mainly modules, i.e.~\emph{Span Extractor}, and~\emph{Mention Classifier}.
    
    \item In~{SpanProto}, we introduce a global boundary matrix to learn the explicit span boundary information. Additionally, we effectively train the model with prototypical learning and margin-based learning to improve the abilities of generalization and adaptation on novel-class entities.
    
    \item Extensive experiments conducted over two widely-used benchmarks illustrate that our method achieves the best performance.
\end{itemize}

\section{Related Work}

In this section, we briefly summarize the related work in various aspects.

\noindent\textbf{Few-shot Learning and Meta Learning.}
Few-shot learning is a challenging problem which aims to learn models that can quickly adapt to different tasks with low-resource labeled data~\cite{Wang2020Generalizing, Huisman2021a}.
A series of typical meta-learning algorithms for few-shot learning consist of optimization-based learning~\cite{Kulkarni2016Domain}, metric-based learning~\cite{Snell2017Prototypical}, and augmentation-based learning~\cite{Wei2019EDA}, etc.
Recently, multiple approaches have been applied in few-shot NLP tasks, such as text classification~\cite{Geng2020Dynamic}, question answering~\cite{Jianing2022KECP} and knowledge base completion~\cite{Sheng2020Adaptive}.
Our {\model} is a typical $N$-way $K$-shot paradigm, which is based on meta learning to make the model better adapt to new domains with little training data available.

\noindent\textbf{Few-shot Named Entity Recognition.} 
Few-shot NER aims to identify and classify the entity type based on low-resource data. 
Recently,~\citet{Ding2020Few} and~\citet{Hou2020Few} provide well-designed few-shot NER benchmarks in a unified $N$-way $K$-shot paradigm. 
A series of approaches typically adopt the metric-based learning method to learn the representations of the entities in the semantic space, i.e. prototypical learning~\cite{Snell2017Prototypical}, margin-based learning~\cite{Levi2021Rethinking} and contrastive learning~\cite{Gao2021SimCSE}. 
Existing approaches can be divided into two kinds, i.e., one-stage~\cite{Snell2017Prototypical, Hou2020Few, Das2022CONTaiNER, Ziyadi2020Example} and two-stage~\cite{Ma2022Decomposed, Wu2022Propose, Shen2021Locate}. Generally, the methods in the kind of one-stage typically categorize the entity type by the token-level metric learning. In contrast, two-stage mainly focuses on two training stages consist of entity span extraction and mention type classification.
\cite{Ma2022Decomposed} is a related work of our paper, which utilizes model-agnostic meta-learning (MAML)~\cite{Finn2017Model} to improve the adaptation ability of the two-stage model.
Different from them, we aim to 1) further improve the performance on detecting and extracting the candidate entity spans, 
and 2) alleviate the bottleneck of false positives in the two-stage few-shot NER system.

\begin{figure*}
\centering
\includegraphics[width=\linewidth]{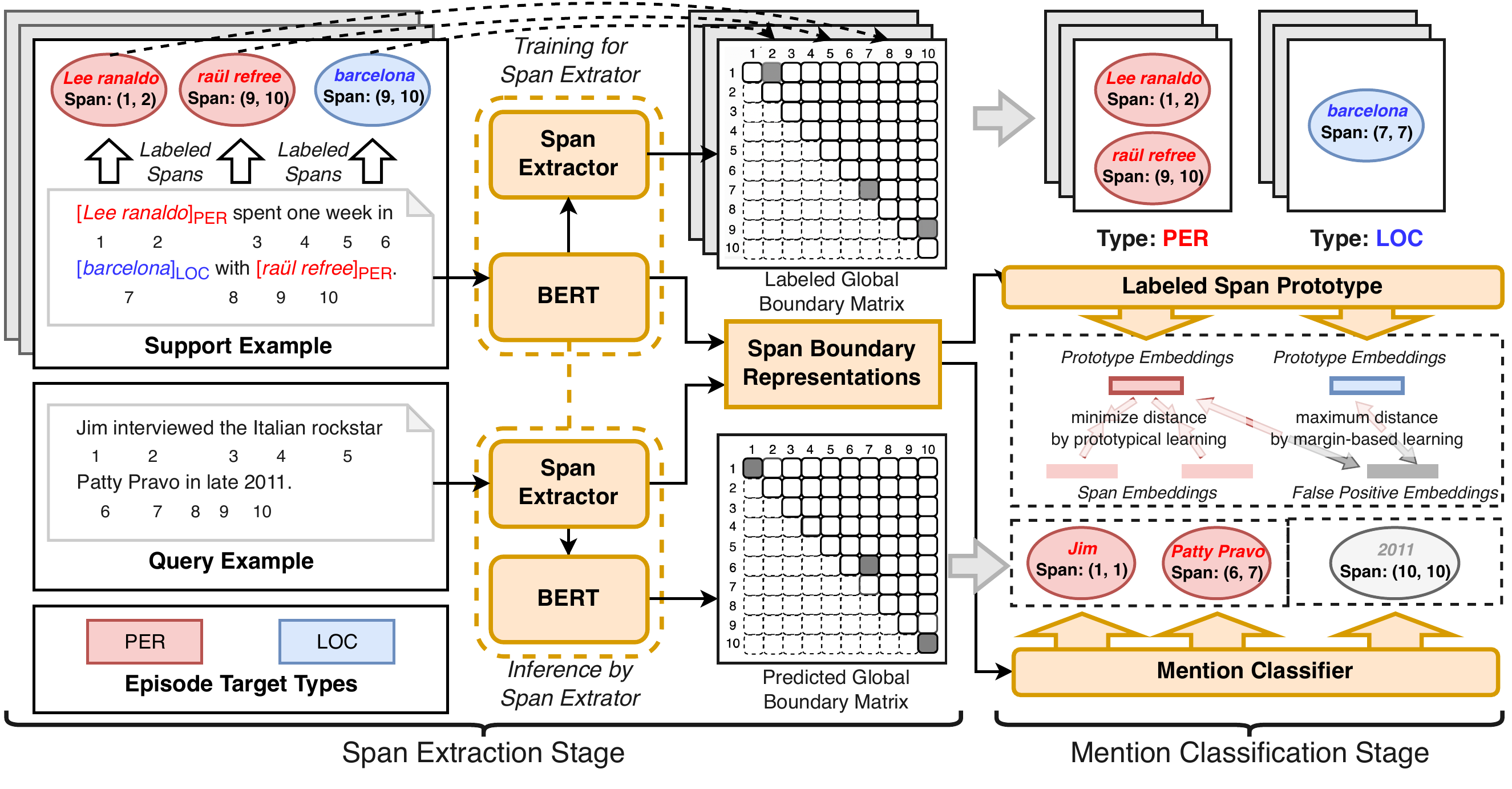}
\caption{The model architecture of {\model}. Given one episode data, we first transform the sequential tags into labeled spans and obtain the corresponding global boundary matrix. We train the span extractor over support examples and predict the spans for query examples. In the classification stage, we introduce prototypical learning and margin-based learning to teach the model to learn better semantic representations. (Best viewed in color.)}
\label{fig:model}
\end{figure*}

\section{Our Proposal: {\model}}

We formally present the notations and the techniques of our proposed {\model}. The model architecture is shown in Figure~\ref{fig:model}.

\subsection{Notations}

Different from
token-wise classification~\cite{Ding2020Few, Hou2020Few}, we define the span-based $N$-way $K$-shot setting for few-shot NER. Suppose that we have a training set $\mathcal{D}_{train}$ and an evaluation set $\mathcal{D}_{eval}$. 
Given one training episode data $\mathcal{E}_{train} = (\mathcal{S}_{train}, \mathcal{Q}_{train}, \mathcal{T}_{train})\in\mathcal{D}_{train}$, where $\mathcal{S}_{train}$ and $\mathcal{Q}_{train}$ denote the support set and the query set, respectively. $\mathcal{T}_{train}$ denotes the entity type set, and $|\mathcal{T}_{train}|=N$. 
For each example $(X, \mathcal{M}, \mathcal{Y})\in\mathcal{S}_{train}\cup\mathcal{Q}_{train}$, $X=\{x_i\}_{i=1}^{L}$ denotes the input sentence with $L$ language tokens. 
$\mathcal{M}=\{(s_j, e_j)\}_{j=1}^{L'}$ represents the mention span set for the sentence $X$, where $s_j, e_j$ denote the start and end position in the sentence for the $j$-th span, and $0\leq s_j\leq e_j\leq L$. $L'$ is the number of spans. 
$\mathcal{Y}=\{y_j\}_{j=1}^{L'}$ is the entity type set
and $y_j\in\mathcal{T}_{train}$ is the label for the $j$-th span $(s_j, e_j)$. 


\subsection{Span Extractor}

The span extractor aims to generate all candidate entity spans. 
Specifically, given one training episode data $\mathcal{E}_{train} = (\mathcal{S}_{train}, \mathcal{Q}_{train}, \mathcal{T}_{train})$ from $\mathcal{D}_{train}$, we use the support example $(X_{s}, \mathcal{M}_{s}, \mathcal{Y}_{s})\in\mathcal{S}_{train}$ to train the span extractor, where $X_{s}=\{x_i\}_{i=1}^{L}$ denotes the input sentence with $L$ tokens, $\mathcal{M}_{s}$ and $\mathcal{Y}_{s}$ are the labeled spans and labeled entity types for $X_{s}$, respectively. 

We first obtain the contextual embeddings by $\mathbf{H} = \mathcal{F}(X_{s})$,
where $\mathcal{F}(\cdot)$ denotes the encoder (e.g. BERT~\cite{Devlin2019BERT}), $\mathbf{H}\in\mathbb{R}^{L\times h}$ is the output representation at the last layer of the encoder, $h$ denotes the embedding size. For each token $x_i\in X_{s}$, the token embedding is denoted as $\mathbf{h}_i\in\mathbb{R}^{h}$.

Rather than detecting the entity mention based on token-wise sequence labeling~\cite{Ma2022Decomposed}, we expect that the span extractor captures the explicit boundary information regardless of the token-wise label dependency. 
We follow the idea of self-attention~\cite{Vaswani2017Attention} which can be used to calculate the attentive correlation for a token pair $(x_i, x_j)$ without any dependencies. Specifically, we first compute the~\emph{query} and~\emph{key} item for each token $x_i$. Formally, we have:
\begin{equation}
\begin{aligned}
\mathbf{q}_i = W_q\mathbf{h}_i + b_q,
~\mathbf{k}_i = W_k\mathbf{h}_i + b_k
\label{eql:key-query}
\end{aligned}
\end{equation}
where $\mathbf{q}_i, \mathbf{k}_i\in\mathbb{R}^{h}$ denote the~\emph{query} and~\emph{key} embeddings, respectively. $W_q, W_k\in\mathbb{R}^{h\times h}$ are the trainable weights and $b_q, b_k\in\mathbb{R}^{h}$ denote the bias. 
Similar to the biaffine decoder~\cite{Yu2020Named}, we design a score function $f(i, j)$ to evaluate the probability of the token pair $(x_i, x_j)$ being an entity span boundary:
\begin{equation}
\begin{aligned}
f(i, j) = \mathbf{q}_{i}^{T}\mathbf{k}_{j} + W_v(\mathbf{h}_{i} + \mathbf{h}_j)
\label{eql:score}
\end{aligned}
\end{equation}
where $W_v\in\mathbb{R}^{h\times h}$ denotes the trainable weight. 
Based on the label spans, we then generate a labeled global boundary matrix for each support sentence:
\begin{equation}
\begin{aligned}
\Omega_{i, j} = \left\{
        \begin{array}{rcl}
        1     & & i\leq j\land(i, j)\in\mathcal{M}_{s}; \\
        0     & & i\leq j\land(i, j)\notin\mathcal{M}_{s}; \\
        -\inf & & i > j; \\
        \end{array} 
    \right.
\label{eql:matrix}
\end{aligned}
\end{equation}
where $\Omega_{i,j}$ is the score of the span $(i, j)$. To explicit learn the span boundary, inspired by~\citet{kexuefm-8373}\footnote{\url{https://spaces.ac.cn/archives/8373.}}, we design a span-based cross-entropy loss to urge the model to learn the boundary information on each training support example:
\begin{equation}
\begin{aligned}
\mathcal{L}_{span} = \log\Bigg(1 + \sum_{1\leq i\leq j\leq L}\exp{\Big((-1)^{\Omega_{i, j}}f(i, j)\Big)}\Bigg)
\label{eql:span-loss}
\end{aligned}
\end{equation}

For each query example $(X_{q}, \mathcal{M}_{q}, \mathcal{Y}_{q})$ in $\mathcal{Q}_{train}$, we can obtain the predicted global boundary matrix $\Omega'$ generated by the span extractor. We follow~\cite{Yu2020Named} to recall the candidate spans which have a higher score than the pre-defined confidence threshold $\theta$. We denote $\hat{\mathcal{M}_q}=\{(s_j, e_j)|\Omega'_{s_j, e_j}\geq \theta\}$ as the predict result. 

\subsection{Mention Classifier}

In the mention classification stage, we introduce a mention classifier to assign the pre-defined entity type for each generated span in the query example.
We leverage prototypical learning to teach the model to learn the semantic representations for each labeled span and enable it to adapt to a new-class domain. To remedy the problem of false positives, we further propose a margin-based learning objective to improve the model performance. 
\subsubsection{Prototypical Learning}
Specifically,  given one episode data $\mathcal{E}_{train} = (\mathcal{S}_{train}, \mathcal{Q}_{train}, \mathcal{T}_{train})$, we can compute the prototype $\mathbf{c}_t\in\mathbb{R}^{h}$ for each class by averaging the representations of all spans in $\mathcal{S}_{train}$ which share the same entity type $t\in\mathcal{T}_{train}$. Formally, we have:
\begin{equation}
\begin{aligned}
\mathbf{c}_t = \frac{1}{K_t}\sum_{(X_s, \mathcal{M}_s, \mathcal{Y}_s)\in\mathcal{S}_{train}\atop}\sum_{(s_j, e_j)\in\mathcal{M}_{s}\atop y_j\in\mathcal{Y}_s}\mathbb{I}( y_j=t)\mathbf{u}_{j}
\label{eql:prototype}
\end{aligned}
\end{equation}
where
\begin{equation}
\begin{aligned}
K_t = \sum_{(X_s, \mathcal{M}_s, \mathcal{Y}_s)\in\mathcal{S}_{train}\atop}\sum_{(s_j, e_j)\in\mathcal{M}_{s}\atop y_j\in\mathcal{Y}_s}\mathbb{I}( y_j=t)
\label{eql:span-num}
\end{aligned}
\end{equation}
represents the span number of the entity type $t$. $\mathbf{u}_{j}\in\mathbb{R}^{h}$ denotes the span boundary representations, and $\mathbf{u}_{j} = \mathbf{h}_{s_j} + \mathbf{h}_{e_j}$. $\mathbb{I}(\cdot)$ is the indicator function.

Note that, the labeled span set $\mathcal{M}_q$ and corresponding type set $\mathcal{Y}_q$ for each query sentence $X_q$ in $\mathcal{Q}_{train}$ is visible during the training stage. Hence, the prototypical learning loss for each sentence can be calculated as:
\begin{equation}
\begin{aligned}
\mathcal{L}_{proto} = \frac{1}{|\mathcal{M}_q|}\sum_{(s_j, e_j)\in\mathcal{M}_q\atop y_j\in\mathcal{Y}_q}-\log p(y_j| s_j, e_j)
\label{eql:proto-loss}
\end{aligned}
\end{equation}
where 
\begin{equation}
\begin{aligned}
p(y_j| s_j, e_j) = \text{SoftMax}(-d(\mathbf{u}_j, \mathbf{c}_{y_j}))
\label{eql:proto-proba}
\end{aligned}
\end{equation}
is the probability distribution with a distance function $d(\cdot, \cdot)$.

\subsubsection{Margin-based Learning}
\label{sec:margin}
Recall that the span set $\hat{\mathcal{M}}_q$ is  extracted by the span extractor for the query sentence $X_q\in\mathcal{Q}_{train}$. We find that some mention spans in $\hat{\mathcal{M}}_q$ do not exist in the ground truth $\mathcal{M}_q$ which can be viewed as false positives. Formally, We denote $\mathcal{M}_q^-=\{(s_j, e_j)|(s_j, e_j)\in\hat{\mathcal{M}}_q\land(s_j, e_j)\notin\mathcal{M}_q\}$ as the set of these false positives. Take Figure~\ref{fig:model} as an example, the span extractor generates four candidate mentions for the query example, i.e., ``Jim'', ``Patty Pravo'', ``Italian'' and ``2011'', where ``Italian'' and ``2011'' are the false positives in this case.

Intuitively, the false positive can be viewed as a special entity mention, which has no type to be assigned in $\mathcal{T}_{train}$, but could be an entity in other episode data. In other words, the real type of this false positive is unknown. Thus, a natural idea is that we can keep it away from all current prototypes in the semantic space. Specifically, we have:
\begin{equation}
\begin{aligned}
\mathcal{L}_{mrg} = & \frac{1}{|\mathcal{T}_{train}||\mathcal{M}_{q}^-|}\sum_{t\in\mathcal{T}_{train}}\sum_{(s_j, e_j)\in\mathcal{M}_{q}^-} \\
& \max(0, r - d(\mathbf{u}_j^-, \mathbf{c}_t))
\label{eql:margin-loss}
\end{aligned}
\end{equation}
where $\mathbf{u}_j^-\in\mathbb{R}^{h}$ denotes the span boundary representations of the false positive span $(s_j, e_j)\in\mathcal{M}_q^-$. Specially, we let $\mathcal{L}_{mrg}$ be 0 if $\mathcal{M}_{q}^-=\varnothing$. $r>0$ is the pre-defined margin value. 
Under the margin-based learning, we can obtain a noise-aware model by pulling away the false positive spans from all prototype regions which can be viewed as the hypersphere with a radius $r$.

\begin{algorithm}[t]
\caption{Training Procedure of {\model}}
\label{alg:train}
\begin{small}
\begin{algorithmic}[1]
\REQUIRE Training data $\mathcal{D}_{train}$, Training total step $T$, Pre-training step of the span extractor $T'<T$.
\FOR {each step $st\in\{1, 2, \cdots, T\}$}
\STATE $\lambda=0$ if $st<T'$ else $\lambda=1$;
\STATE Sample $(\mathcal{S}_{train}, \mathcal{Q}_{train}, \mathcal{T}_{train})$ from $\mathcal{D}_{train}$;
\STATE $\mathcal{L}_{span\_eps}=\mathcal{L}_{proto\_eps}=\mathcal{L}_{mrg\_eps}=0.$
\FOR {$(X_s, \mathcal{M}_s, \mathcal{Y}_s)\in\mathcal{S}_{train}$}
\STATE Compute $\mathcal{L}_{span}$ for the span extractor in Eq.~\ref{eql:span-loss};
\STATE $\mathcal{L}_{span\_eps} = \mathcal{L}_{span\_eps} + \mathcal{L}_{span}$;
\ENDFOR
\STATE Get the prototype for each type $t\in\mathcal{T}_{train}$ in Eq.~\ref{eql:prototype};
\FOR {$(X_q, \mathcal{M}_q, \mathcal{Y}_q)\in\mathcal{Q}_{train}$}
\STATE Based on $\mathcal{M}_q$ and $\mathcal{Y}_q$, calculate prototypical learning loss $\mathcal{L}_{proto}$ in Eq.~\ref{eql:proto-loss};
\STATE Obtain the false positives set $\mathcal{M}_q^-$,
and calculate margin-based learning loss $\mathcal{L}_{mrg}$ in Eq.~\ref{eql:margin-loss};
\STATE $\mathcal{L}_{proto\_eps} = \mathcal{L}_{proto\_eps} + \mathcal{L}_{proto}$;
\STATE $\mathcal{L}_{mrg\_eps} = \mathcal{L}_{mrg\_eps} + \mathcal{L}_{mrg}$;
\ENDFOR
\STATE Obtain the overall loss by $\mathcal{L}=\frac{1}{|\mathcal{S}_{train}|}\mathcal{L}_{span\_eps} + \frac{\lambda}{|\mathcal{Q}_{train}|}(\mathcal{L}_{proto\_eps} + \mathcal{L}_{mrg\_eps})$;
\STATE Update the model by backpropagation to reduce $\mathcal{L}$;
\ENDFOR
\RETURN The trained {\model} model.
\end{algorithmic}
\end{small}
\end{algorithm}

\subsection{Training and Evaluation of {\model}}

We provide a brief description of the training and evaluate the algorithm for our framework. During the training stage, the procedure can be found in Algorithm~\ref{alg:train}. 
For each training step, we randomly sample one episode data from $\mathcal{D}_{train}$, and then enumerate each support example to obtain the span-based loss (Algorithm~\ref{alg:train}, Line 3-8).
For the support set, we obtain the span boundary representations, and then calculate the prototype for each target type (Algorithm~\ref{alg:train}, Line 9). 
Further, we leverage the span extractor to perform model inference on each query example  to recall multiple candidate spans, and then compute the prototypical learning loss and margin-based learning loss (Algorithm~\ref{alg:train}, Line 10-15). 
For the model training, the total training steps denote as $T$. We first pre-train the span extractor for $T'(T'<T)$ steps, and then both of the span extractor and mention classifier are jointly trained with three objective losses (Algorithm~\ref{alg:train}, Line 16-17).

During the model evaluation, given one episode data $\mathcal{E}_{eval}=(\mathcal{S}_{eval}, \mathcal{Q}_{eval}, \mathcal{T}_{eval})\in\mathcal{D}_{eval}$. We first calculate the prototype representations based on the support set $\mathcal{S}_{eval}$. 
Then, for each query sentence in $\mathcal{Q}_{eval}$, we can utilize the the span extractor to extract all candidate spans $\hat{\mathcal{M}}_q$. 
In the mention classification stage, we calculate the distance between the extracted span and each prototype, and select the type of the nearest prototype as the result. 
Note that the ground truth in $\mathcal{D}_{eval}$ is invisible, to recognize the false positives, we remove all spans whose distances between them and all prototypes are larger than $r$.

\begin{table}
\centering
\resizebox{\linewidth}{!}{
\begin{small}
\begin{tabular}{l | cccc}
\toprule
\bf Datasets &\bf  Domain &\bf  \#Types &\bf  \#Sentences & \bf \#Entities \\
\midrule
Few-NERD & General & 66 & 188.2k & 491.7k \\
OntoNotes & General & 18 & 103.8k & 161.8k \\
CoNLL-03 & News & 4 & 22.1k & 35.1k \\
GUM & Wiki & 11 & 3.5k & 6.1k \\
WNUT-17 & Social & 6 & 4.7k & 3.1k \\
\bottomrule
\end{tabular}
\end{small}
}
\caption{The statistics of each source dataset.}
\label{tab:datasets}
\end{table}

\section{Experiments}

\subsection{Datasets and Baselines}
We choose two widely used $N$-way $K$-shot based benchmarks to evaluate our {\model}, including Few-NERD~\cite{Ding2020Few}~\footnote{\url{https://github.com/thunlp/Few-NERD}.} and CrossNER~\cite{Hou2020Few}. Specifically, \textbf{Few-NERD} is annotated with 8 coarse-grained and 66 fine-grained entity types, which consists of two few-shot settings, i.e. Intra, and Inter. In the \textbf{Intra} setting, all entities in the training set, development set, and testing set belong to different coarse-grained types. In contrast, in the \textbf{Inter} setting, only the fine-grained entity types are mutually disjoint in different datasets. To make a fair comparison, we utilize the processed episode data released by~\citet{Ding2020Few}. \textbf{CrossNER} consists of four different NER domains, such as OntoNotes 5.0~\cite{weischedel2013ontonotes}, CoNLL-03~\cite{Sang2003Introduction}, GUM~\cite{Zeldes2017The} and WNUT-17~\cite{Derczynski2017Results}. During training, we randomly select two of them as the training set, and the remaining two others are used for the development set and the testing set. The final results are derived from the testing set. We follow~\cite{Ma2022Decomposed} to utilize the generated episode data provided by~\citet{Hou2020Few}~\footnote{\url{https://atmahou.github.io/attachments/ACL2020data.zip}.}.

\begin{table*}[t]
    \centering
    \setlength{\tabcolsep}{1mm}
    \resizebox{\linewidth}{!}{
    \begin{tabular}{l l | ccccc | ccccc}
    \toprule
        \multirow{3}{*}{\textbf{Paradigms}} &
        \multirow{3}{*}{\textbf{Models}} &
        \multicolumn{5}{c|}{\textbf{Intra}} & \multicolumn{5}{c}{\textbf{Inter}}\\
        \cmidrule(lr){3-7} \cmidrule(lr){8-12}
        & & \multicolumn{2}{c}{\textbf{1$\sim$2-shot}} & \multicolumn{2}{c}{\textbf{5$\sim$10-shot}} & \multirow{2}{*}{Avg.} & \multicolumn{2}{c}{\textbf{1$\sim$2-shot}} & \multicolumn{2}{c}{\textbf{5$\sim$10-shot}} & \multirow{2}{*}{Avg.}\\
         & & 5 way & 10 way & 5 way & 10 way & & 5 way & 10 way & 5 way & 10 way & \\
        \midrule
        \multirow{4}{*}{\emph{One-stage}} & ProtoBERT$^{\dag}$ & 23.45\small\small{\textpm0.92} & 19.76\small{\textpm0.59} & 41.93\small{\textpm0.55} & 34.61\small{\textpm0.59} & 29.94 &  44.44\small{\textpm0.11} & 39.09\small{\textpm0.87} & 58.80\small{\textpm1.42} & 53.97\small{\textpm0.38} & 49.08 \\
        & NNShot$^{\dag}$ & 31.01\small{\textpm1.21} & 21.88\small{\textpm0.23} & 35.74\small{\textpm2.36} & 27.67\small{\textpm1.06} & 29.08 & 54.29\small{\textpm0.40} & 46.98\small{\textpm1.96} & 50.56\small{\textpm3.33} & 50.00\small{\textpm0.36} & 50.46 \\
        & StructShot$^{\dag}$ & 35.92\small{\textpm0.69} & 25.38\small{\textpm0.84} & 38.83\small{\textpm1.72} & 26.39\small{\textpm2.59} & 31.63 & 57.33\small{\textpm0.53} & 49.46\small{\textpm0.53} & 57.16\small{\textpm2.09} & 49.39\small{\textpm1.77} & 53.34 \\
        & CONTaiNER$^{\ddag}$ & 40.43 & 33.84 & 53.70 & 47.49 & 43.87 & 55.95 & 48.35 & 61.83 & 57.12 & 55.81 \\
        \midrule
        \multirow{3}{*}{\emph{Two-stage}} & 
        ESD & 41.44\small{\textpm1.16} & 32.29\small{\textpm1.10} & 50.68\small{\textpm0.94} & 42.92\small{\textpm0.75} & 41.83 &  66.46\small{\textpm0.49} & 59.95\small{\textpm0.69} & 74.14\small{\textpm0.80} & 67.91\small{\textpm1.41} & 67.12 \\
        & DecomMeta & 52.04\small{\textpm0.44} & 43.50\small{\textpm0.59} & 63.23\small{\textpm0.45} & 56.84\small{\textpm0.14} & 53.90 & 68.77\small{\textpm0.24} & 63.26\small{\textpm0.40} & 71.62\small{\textpm0.16} & 68.32\small{\textpm0.10} & 67.99 \\
        & \textbf{\model} & \textbf{54.49\small{\textpm0.39}} & \textbf{45.39\small{\textpm0.72}} & \textbf{65.89\small{\textpm0.82}} & \textbf{59.37\small{\textpm0.47}} & \textbf{56.29} & \textbf{73.36\small{\textpm0.18}} & \textbf{66.26\small{\textpm0.33}} & \textbf{75.19\small{\textpm0.77}} & \textbf{70.39\small{\textpm0.63}} & \textbf{71.30} \\
        \bottomrule
    \end{tabular}
    }
    \caption{F1 scores with standard deviations on Few-NERD for both inter and intra settings. $^{\dag}$ denotes the results reported in~\citet{Das2022CONTaiNER}. $^{\ddag}$ is taken from~\citet{Das2022CONTaiNER} with no standard deviations reported.}
    \label{tab:fewnerd}
\end{table*}

\begin{table*}[t]
    \centering
    \setlength{\tabcolsep}{1mm}
    \resizebox{\linewidth}{!}{
    \begin{tabular}{ll | ccccc | ccccc}
    \toprule
        \multirow{2}{*}{\textbf{Paradigms}} &
        \multirow{2}{*}{\textbf{Models}} &
        \multicolumn{5}{c|}{\textbf{1-shot}} & \multicolumn{5}{c}{\textbf{5-shot}}\\
        \cmidrule(lr){3-7} \cmidrule(lr){8-12}

        & & CONLL-03 & GUM & WNUT-17 & OntoNotes & Avg. & CONLL-03 & GUM & WNUT-17 & OntoNotes & Avg. \\
        \midrule
        \multirow{3}{*}{\emph{One-stage}} & Matching Network$^{\ddag}$ & 19.50\small{\textpm0.35} & 4.73\small{\textpm0.16} & 17.23\small{\textpm2.75} & 15.06\small{\textpm1.61} & 14.13 & 19.85\small{\textpm0.74} & 5.58\small{\textpm0.23} & 6.61\small{\textpm1.75} & 8.08\small{\textpm0.47} & 10.03 \\
        & ProtoBERT$^{\ddag}$ & 32.49\small{\textpm2.01} & 3.89\small{\textpm0.24} & 10.68\small{\textpm1.40} & 6.67\small{\textpm0.46} & 13.43 &  50.06\small{\textpm1.57} & 9.54\small{\textpm0.44} & 17.26\small{\textpm2.65} & 13.59\small{\textpm1.61} & 22.61 \\
        & L-TapNet+CDT & 44.30\small{\textpm3.15} & 12.04\small{\textpm0.65} & 20.80\small{\textpm1.06} & 15.17\small{\textpm1.25} & 23.08 & 45.35\small{\textpm2.67} & 11.65\small{\textpm2.34} & 23.30\small{\textpm2.80} & 20.95\small{\textpm2.81} & 25.31 \\
        \midrule
        \multirow{2}{*}{\emph{Two-stage}} & DecomMeta & 46.09\small{\textpm0.44} & 17.54\small{\textpm0.98} & 25.14\small{\textpm0.24} & 34.13\small{\textpm0.92} & 30.73 & 58.18\small{\textpm0.87} & 31.36\small{\textpm0.91} & 31.02\small{\textpm1.28} & 45.55\small{\textpm0.90} & 41.53 \\
        & \textbf{\model} & \textbf{47.70\small{\textpm0.49}} & \textbf{19.92\small{\textpm0.53}} & \textbf{28.31\small{\textpm0.61}} & \textbf{36.41\small{\textpm0.73}} & \textbf{33.09} & \textbf{61.88\small{\textpm0.83}} & \textbf{35.12\small{\textpm0.88}} & \textbf{33.94\small{\textpm0.50}} & \textbf{48.21\small{\textpm0.89}} & \textbf{44.79}\\
        \bottomrule
    \end{tabular}
    }
    \caption{F1 scores with standard deviations on CrossNER. $^{\ddag}$ denotes the results reported in~\citet{Hou2020Few}.}
    \label{tab:crossner}
\end{table*}

For the baselines, we choose multiple strong approaches from the paradigms of~\emph{one-stage} and~\emph{two-stage}. Concretely, the one-stage paradigm consists of ProtoBERT~\cite{Snell2017Prototypical}, Matching Network~\cite{Vinyals2016Matching}, StructShot~\cite{Yang2020Simple}, NNShot~\cite{Yang2020Simple}, CONTaiNER~\cite{Das2022CONTaiNER} and L-TapNet+CDT~\cite{Hou2020Few}. 
The two-stage paradigm includes ESD~\cite{Wang2021An} and DecomMeta~\cite{Ma2022Decomposed}.

\subsection{Implementation Details}
We choose BERT-base-uncased~\cite{Devlin2019BERT} from HuggingFace\footnote{\url{https://huggingface.co/transformers}.} as the default pre-trained encoder $\mathcal{F}$. The max sequence length we set is 128. We choose AdamW as the optimizer with a warm up rate of 0.1.
The training steps $T$ and $T'$ are set as 2000 and 200, respectively. 
We use the grid search to find the best hyper-parameters for each benchmark~\footnote{The details are shown in Appendix~\ref{app:grid-search}}. 
As a result, the threshold $\theta$ and the margin $r$ are set as 0.8 and 3.0, respectively.
We choose five random seeds from \{12, 21, 42, 87, 100\} and report the averaged results with standard deviations.
We implement our model by Pytorch 1.8, and train the model with 8 V100-32G GPUs.

\subsection{Main Results}
Table~\ref{tab:fewnerd} and~\ref{tab:crossner} illustrate the main results of our proposal compared with other baselines. We thus make the following observations: 1) Our proposed~{\model} achieves the best performance and outperforms the state-of-the-art baselines with a large margin. Compared with DecomMeta~\cite{Ma2022Decomposed}, the overall averaged results over Few-NERD Intra and Few-NERD Inter are improved by 2.39\% and 3.31\%, respectively. Likewise, we also have more than 2.0\% advantages on CrossNER. 
2) All the methods in the two-stage paradigm perform better than those one-stage approaches, which indicates the merit of the span-based approach for the task of few-shot NER.
3) In Few-NERD, the overall performance of the Intra scenario is lower than Inter. This phenomenon reflects that Intra is more challenging than Inter where the coarse-grained types are different in training/development/testing set. Despite this, we still obtain satisfying effectiveness. Results suggest that our method can adapt to a new domain in which the coarse-grained and fine-grained entity types are both unseen.

\subsection{Ablation Study}
We conduct an ablation study to investigate the characteristics of the main components in~{\model}. We implement the following list of variants of~{\model} for the experiments. 
1) w/o. Span Extractor: we remove the span extractor, and train the model with a conventional token-wise prototypical network~\footnote{When removing the span extractor, the margin-based learning will not work because no spans can be recalled.}. 
2) w/o. Mention Classifier: we remove all techniques in the mention classifier. To classify the span, we directly leverage the K-Means algorithm. 
3) w/o. Margin-based Learning: we only remove the margin-based learning objective. More details of these variants are shown in Appendix~\ref{app:variants}. 

\begin{table}
\centering
\resizebox{\linewidth}{!}{
\begin{tabular}{l | cccc}
\toprule
\multirow{2}{*}{\textbf{Methods}} &  \multicolumn{2}{c}{\textbf{Few-NERD}}  &  \multicolumn{2}{c}{\textbf{CrossNER}} \\
& Intra & Inter & 1-shot & 5-shot \\
\midrule
\textbf{\model} & \bf 56.29 & \bf 71.30 & \bf 33.09 & \bf 44.79 \\
\midrule
w/o. Span Extractor & 29.24 & 49.08 & 13.43 & 22.61 \\
w/o. Mention Classifier & 18.41 & 26.36 & 10.53 & 6.08 \\
w/o. Margin-based Learning & 54.29 & 71.37 & 30.88 & 43.37 \\
\bottomrule
\end{tabular}
}
\caption{The ablation study results (averaged F1 score \%) for Few-NERD and CrossNER. Detail results are listed in Appendix~\ref{app:ablation}.}
\label{tab:ablation}
\end{table}

\begin{table}
\centering
\begin{small}
\begin{tabular}{l | cccc}
\toprule
\multirow{2}{*}{\textbf{Methods}} &  \multicolumn{2}{c}{\textbf{Few-NERD}}  &  \multicolumn{2}{c}{\textbf{CrossNER}} \\
& Intra & Inter & 1-shot & 5-shot \\
\midrule
ESD & 70.56 & 70.99 & - & - \\
DecomMeta & 76.11 & 76.48 & 46.53 & 54.58 \\
\textbf{\model} & \bf 84.02 & \bf 84.55 & \bf 63.75 & \bf 63.51 \\
\bottomrule
\end{tabular}
\end{small}
\caption{The averaged performance (F1 score \%) in the span extraction stage over Few-NERD and CrossNER.}
\label{tab:span_comparison}
\end{table}

As shown in Table~\ref{tab:ablation}, the results demonstrate that 
1) the performance of~{\model} drops when removing each component, which shows the significance of all components.
2) When removing the span extractor, the averaged F1 scores are decreased by 
more than 20\%, indicating that the span extractor which bypasses the issues of multiple non-entity classes and token-level label dependency does make a contribution to the model performance.
3) {\model} outperforms w/o. Margin-based Learning, which demonstrates that a model jointly trained by prototypical learning and margin-based learning objectives can mitigate the problem of false positives.

\subsection{Performance of the Span Extractor}
To further analyze how the span extractor contributes to the few-shot NER. Specifically, we aim to solve the following research questions about the span extractor: 1) \textbf{RQ1}: Whether the proposed span extractor is
better than previous methods? 2) \textbf{RQ2}: How does the model learn explicit span boundary information? 3) \textbf{RQ3}: How do the hyper-parameters $T'$ and $\theta$ affect the model performance?

\begin{figure}
\centering
\includegraphics[width=0.95\linewidth]{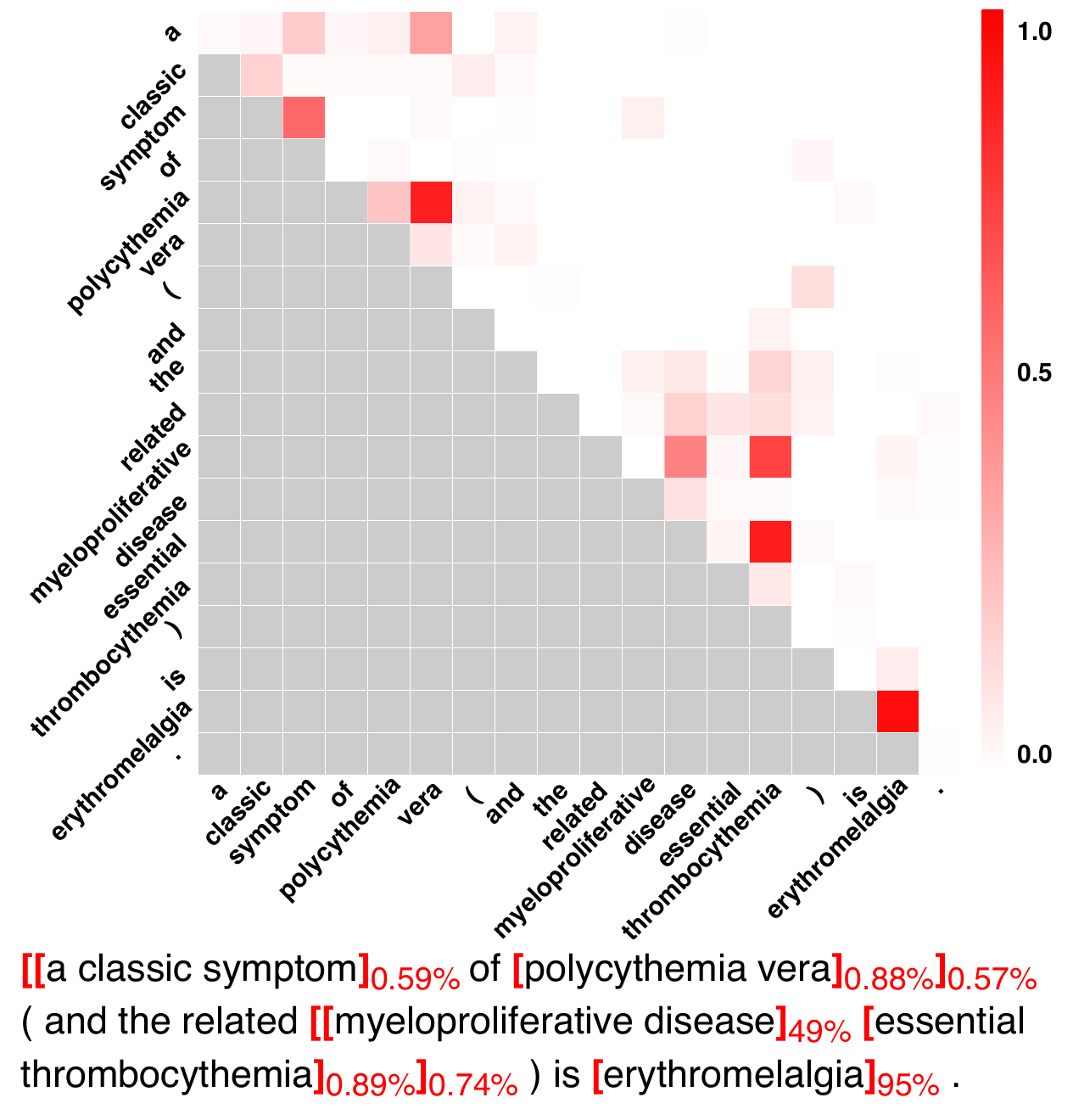}
\caption{The case visualization of the global boundary matrix. The span could be an entity if the corresponding color is dark. (Best viewed in color.)}
\label{fig:span_boundary_case}
\end{figure}

\noindent\textbf{Comparison with Baselines.}
We first perform a comparison between our proposal and previous methods. Specifically, we choose two strong baselines: 1) ESD~\cite{Wang2021An} which leverages a sliding window to recall all candidate spans, and 2) DecomMeta~\cite{Ma2022Decomposed} which leverages the sequence labeling method with ``BIOE'' rules to detect the spans.
From Table~\ref{tab:span_comparison}, we see that our span extractor outperforms other baselines by a large margin, which indicates that our method can produce more accurate predictions for span extraction. 

\noindent\textbf{Case Study for the Span Extractor.}
To explore how the model learns the span boundary, we randomly select one query sentence from Few-NERD and obtain the predicted global boundary matrix. As shown in Figure~\ref{fig:span_boundary_case}, our span extractor successfully recognizes the explicit boundary for the input sentence. 
In addition, we find that there existing nested entity mentions in few-shot NER, which share the same overlap sub-sequence. For example, ``myeloproliferative disease'', ``essential thrombocythemia'', and ``myeloproliferative disease essential thrombocythemia'' are the nested entities. Based on the global boundary matrix, our approach can be extended to few-shot nested NER without difficulty, which is ignored by previous works.

\begin{figure}[t]
\centering
\begin{tabular}{ll}
\begin{minipage}[t]{0.48\linewidth}
    \includegraphics[width = 1\linewidth]{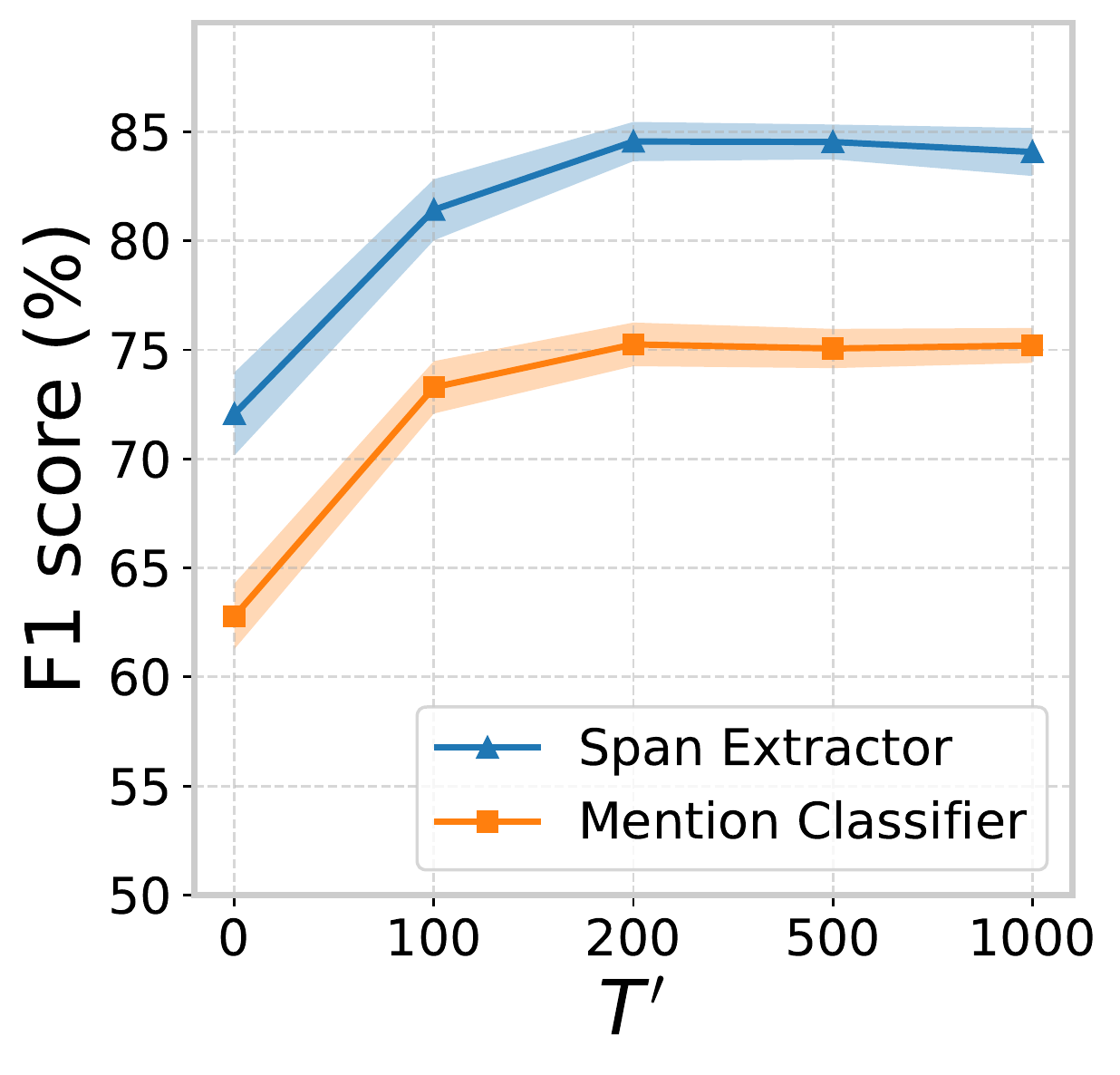}
\end{minipage}
\begin{minipage}[t]{0.48\linewidth}
    \includegraphics[width = 1\linewidth]{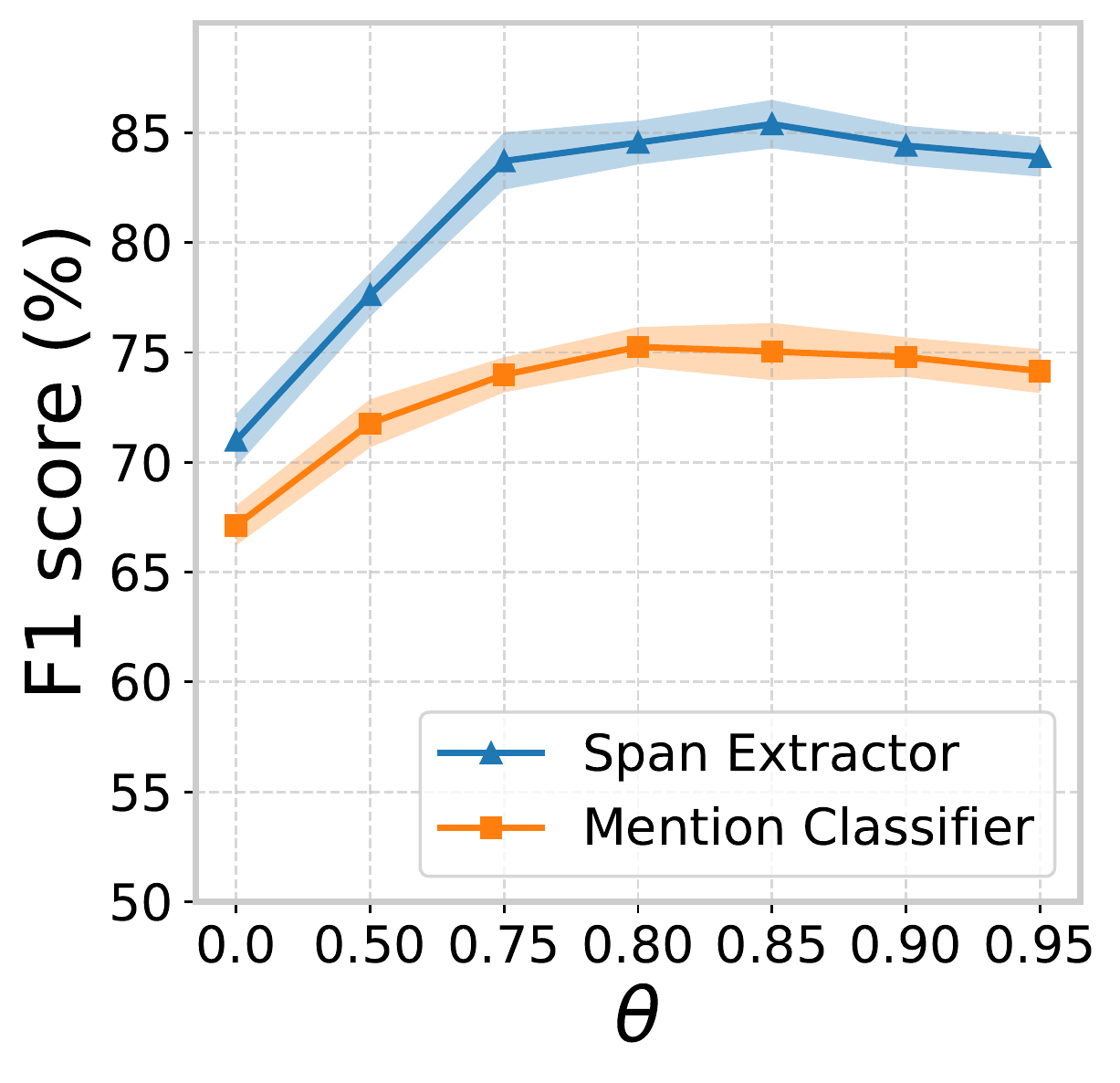}
\end{minipage}
\end{tabular}
\caption{Hyper-parameter effectiveness (averaged F1 score \%) of $T'$ and $\theta$ over Few-NERD Inter. We report the F1 curve for both span extractor and mention classifier. $\theta=0.0$ denotes to recall all mention spans, which is equal to the method of ESD~\cite{Wang2021An}.}
\label{fig:hyper-parameter1}
\end{figure}

\noindent\textbf{Effectiveness of Hyper-parameters.}
We aim to investigate the influence of two hyper-parameters $T'$ and $\theta$, where $T'$ is the pre-training step number of the span extractor, and $\theta$ denotes the confidence threshold for predicting candidate spans. We conduct experiments over Few-NERD Inter. As shown in Figure~\ref{fig:hyper-parameter1}. We can draw the following suggestions:
1) For the parameter $T'$, we find that pre-training the span extractor for $T'=200$ steps does improve the performance for both the span extractor and the mention classifier. When $T'>200$, the F1 scores will decrease due to the over-fitting problem.
2) For the parameter $\theta$, we observe that the overall F1 score increases
when increasing the threshold $\theta$, and we achieve the best performance when $\theta\in[0.8, 0.85]$.

\begin{table}
\centering
\begin{small}
\begin{tabular}{l | cccc}
\toprule
\multirow{2}{*}{\textbf{Margin $r$}} &  \multicolumn{2}{c}{\textbf{Few-NERD}}  &  \multicolumn{2}{c}{\textbf{CrossNER}} \\
& Intra & Inter & 1-shot & 5-shot \\
\midrule
$r=1$ & 51.24 & 61.32 & 26.82 & 39.35 \\
$r=2$ & 55.19 & 69.80 & \bf 33.53 & 44.49 \\
$r=3$ & \bf 56.29 & \bf 71.30 & 33.09 & \bf 44.79 \\
$r=4$ & 55.11 & 71.03 & 32.70 & 43.42 \\
$r=5$ & 53.24 & 70.61 & 31.59 & 43.02 \\
$r=6$ & 52.30 & 68.18 & 30.66 & 42.18 \\
\bottomrule
\end{tabular}
\end{small}
\caption{The parameter analysis of the margin $r$. }
\label{tab:hyper-parameter2}
\end{table}

\subsection{Effectiveness of the Mention Classifier}
In this section, we specifically explore the effectiveness of the mention classifier by answering two questions. 1) \textbf{RQ4}: How does the hyper-parameter $r$ affects the model performance? and 2) \textbf{RQ5}: How does the mention classifier adjust the semantic representations of each entity span?

\begin{figure}
\centering
\includegraphics[width=0.8\linewidth]{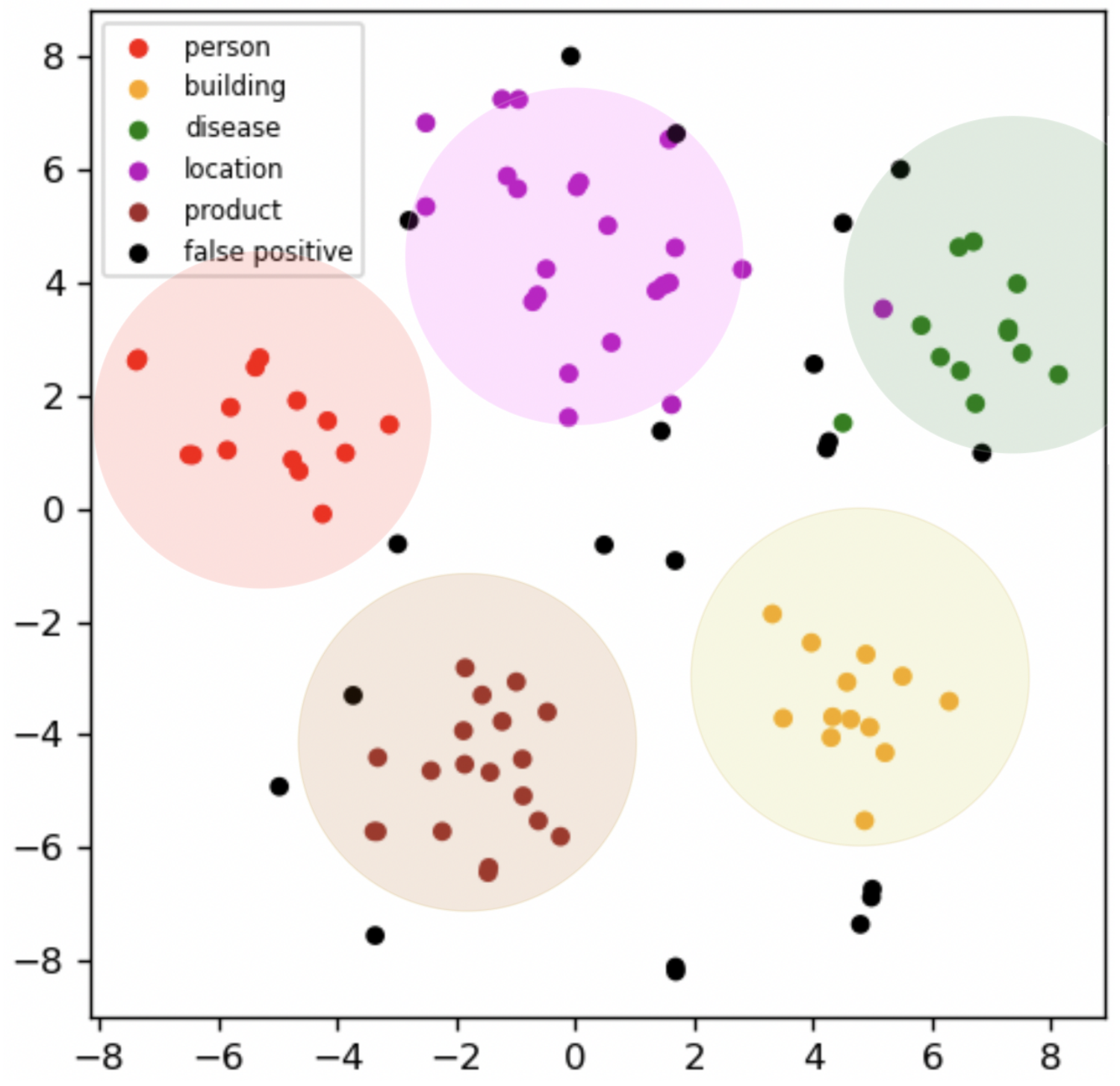}
\caption{The t-SNE visualization of the span representations with 5-way 5$\sim$10-shot episode data from Few-NERD Inter. The points with different colors denote the entity span with different types. The circle represents the prototype region. (Best viewed in color.)}
\label{fig:visualization}
\end{figure}

\noindent\textbf{Effectiveness of the Hyper-parameter.}
We further perform a hyper-parameter study on the margin value $r$, which aims to separate the false positives via the margin-based learning objective. We conduct the experiments over Few-NERD and CrossNER, and the report the averaged F1 scores in Table~\ref{tab:hyper-parameter2}. 
Through the experimental results, we find the best value is in $2\leq r\leq 3$. Specifically, when the margin value is larger than 3, the performance is similar to the variants~{\model} (w/o. Margin-based Learning). In other words, it is hard to detect false positives. In addition, we find that the performance declines a lot when $r<2$. We think that the general distance between the prototype and the true positive span is larger than 2. If the margin value is very small, more and more true positives could be viewed as the noises and be discarded.

\noindent\textbf{Visualizations.}
We end this section with an investigation of how the~{\model} learns the representations in the semantic space. We randomly choose a 5-way 5$\sim$10-shot episode data from Few-NERD Inter, and then obtain the visualization by t-SNE~\cite{van2008visualizing} toolkit. As shown in Figure~\ref{fig:visualization}, our method can cluster the span representations into the corresponding type prototype region, which can be viewed as a circle with a radius of 2. In addition, owing to the proposed margin-based learning, we observe that most of the false positives can be separated successfully.

\begin{table}
\centering
\begin{small}
\begin{tabular}{l | ccc}
\toprule
\bf Methods & \bf F1 & \bf FP-Span & \bf FP-Type  \\
\midrule
ProtoBERT & 44.44 & 86.70 & 13.30 \\
NNShot & 54.29 & 84.70 & 15.30 \\
StructShot & 57.33 & 80.00 & 20.00 \\
ESD & 66.46 & 72.80 & 27.20 \\
DecomMeta & 76.11 & 76.48 & 46.53 \\
\textbf{\model} & \bf 73.36 & \bf 61.40 & \bf 10.90 \\
\bottomrule
\end{tabular}
\end{small}
\caption{Error analysis (\%) of 5-way 1~2-shot on FewNERD-INTER. ``FP-Span'' denotes extracted entities with the wrong span boundary, and ``FP-Type'' represents extracted entities with the right span boundary but the wrong entity type.}
\label{tab:error_analysis}
\end{table}

\subsection{Error Analysis}
Finally, we follow~\citet{Wang2021An} to conduct an error analysis into two types, i.e. ``FP-Span'' and ``FP-Type''. As shown in Table~\ref{tab:error_analysis}, our~{\model} outperforms other strong baselines and has much less false positive prediction errors.
Specifically, we achieve 61.40\% of ``FP-Span'' and the result reduces by more than 10\%, which demonstrates the effectiveness of the span extractor. Meanwhile, we also obtain the lowest error rate of ``FP-Type'', owing to the introduction of the margin-based learning objective in the mention classifier.

\section{Conclusion}
We propose a span-based prototypical network ({\model}) for few-shot NER, which is a~\emph{two-stage} approach includes span extraction and mention classification. To improve the performance of the span extraction, we introduce a global boundary matrix to urge the model to learn explicit boundary information regardless of token-wise label dependency. We further utilize prototypical learning to adjust the span representations and solve the issue of false positives by the margin-based learning objective. Extensive experiments demonstrate that our framework consistently outperforms strong baselines. In the future, we will extend our framework to other NLP tasks, such as slot filling, Part-Of-Speech (POS) tagging, etc.

\section*{Limitations}
Our proposed span extractor is based on the global boundary matrix, which could consume more memory than the conventional sequence labeling methods. This drives us to further improve the overall space efficiency of the framework.
In addition, we only focus on the few-shot NER in a $N$-way $K$-shot settings in this paper. But, we think it is possible to extend our work to other NER scenarios, such as transfer learning, semi-supervised learning or full-data supervised learning. We also leave them as our future research.

\section*{Ethical Considerations}

Our contribution in this work is fully methodological, namely a Span-based Prototypical Network ({\model}) to boost the performance of the few-shot NER. Hence, there are no direct negative social impacts of this contribution. 

\section*{Acknowledgments}
\label{sec:acknowledge}
This work has been supported by the National Natural Science Foundation of China under Grant No. U1911203, 
Alibaba Group through the Alibaba Innovation Research Program, 
the National Natural Science Foundation of China under Grant No. 61877018,
the Research Project of Shanghai Science and Technology Commission (20dz2260300) and The Fundamental Research Funds for the Central Universities.


\bibliography{anthology,custom}
\bibliographystyle{acl_natbib}

\appendix

\begin{table*}[t]
    \centering
    \setlength{\tabcolsep}{1mm}
    \resizebox{\linewidth}{!}{
    \begin{tabular}{l | ccccc | ccccc}
    \toprule
    \multirow{3}{*}{\textbf{Models}} &
    \multicolumn{5}{c|}{\textbf{Intra}} & \multicolumn{5}{c}{\textbf{Inter}}\\
    \cmidrule(lr){2-6} \cmidrule(lr){7-11}
    & \multicolumn{2}{c}{\textbf{1$\sim$2-shot}} & \multicolumn{2}{c}{\textbf{5$\sim$10-shot}} & \multirow{2}{*}{Avg.} & \multicolumn{2}{c}{\textbf{1$\sim$2-shot}} & \multicolumn{2}{c}{\textbf{5$\sim$10-shot}} & \multirow{2}{*}{Avg.}\\
    & 5 way & 10 way & 5 way & 10 way & & 5 way & 10 way & 5 way & 10 way & \\
    \midrule
    \textbf{\model} & \textbf{54.49\small{\textpm0.39}} & \textbf{45.39\small{\textpm0.72}} & \textbf{65.89\small{\textpm0.82}} & \textbf{59.37\small{\textpm0.47}} & \textbf{56.29} & \textbf{73.36\small{\textpm0.18}} & \textbf{66.26\small{\textpm0.33}} & \textbf{75.19\small{\textpm0.77}} & \textbf{70.39\small{\textpm0.63}} & \textbf{71.30} \\

    w/o. Span Extractor & 23.10\small{\textpm0.37} & 21.63\small{\textpm0.29} & 37.91\small{\textpm0.44} & 34.32\small{\textpm0.44} & 29.24 &
    45.17\small{\textpm0.25} & 36.18\small{\textpm0.35} & 59.52\small{\textpm1.0} & 55.45\small{\textpm0.90} & 49.08 \\

    w/o. Mention Classifier & 14.02\small{\textpm0.25} & 11.33\small{\textpm0.33} & 31.20\small{\textpm0.75} & 17.09\small{\textpm0.20} & 18.41 & 25.40\small{\textpm0.22} & 19.77\small{\textpm0.36} & 26.88\small{\textpm0.41} & 33.39\small{\textpm0.50} & 26.36 \\

    w/o. Margin-based Learning & 51.92\small{\textpm0.40} & 40.32\small{\textpm0.52} & 68.10\small{\textpm0.88} & 56.82\small{\textpm0.19} & 54.29 & 68.07\small{\textpm0.22} & 62.52\small{\textpm0.30} & 79.10\small{\textpm0.35} & 75.79\small{\textpm0.33} & 71.37 \\
    \bottomrule
    \end{tabular}
    }
    \caption{F1 scores with standard deviations on Few-NERD for both inter and intra settings. $^{\dag}$ denotes the results reported in~\citet{Das2022CONTaiNER}. $^{\ddag}$ is taken from~\citet{Das2022CONTaiNER} with no standard deviations reported.}
    \label{tab:ablation-1}
\end{table*}

\begin{table*}[t]
    \centering
    \setlength{\tabcolsep}{1mm}
    \resizebox{\linewidth}{!}{
    \begin{tabular}{l | ccccc | ccccc}
    \toprule
    \multirow{2}{*}{\textbf{Models}} &
    \multicolumn{5}{c|}{\textbf{1-shot}} & \multicolumn{5}{c}{\textbf{5-shot}}\\
    \cmidrule(lr){2-6} \cmidrule(lr){7-11}

    & CONLL-03 & GUM & WNUT-17 & OntoNotes & Avg. & CONLL-03 & GUM & WNUT-17 & OntoNotes & Avg. \\
    \midrule
    \textbf{\model} & \textbf{47.70\small{\textpm0.49}} & \textbf{19.92\small{\textpm0.53}} & \textbf{28.31\small{\textpm0.61}} & \textbf{36.41\small{\textpm0.73}} & \textbf{33.09} & \textbf{61.88\small{\textpm0.83}} & \textbf{35.12\small{\textpm0.88}} & \textbf{33.94\small{\textpm0.50}} & \textbf{48.21\small{\textpm0.89}} & \textbf{44.79}\\
    
    w/o. Span Extractor & 9.10\small{\textpm0.37} & 11.13\small{\textpm0.21} & 17.28\small{\textpm0.44} & 16.21\small{\textpm0.40} & 13.43 &
    25.63\small{\textpm0.23} & 21.76\small{\textpm0.39} & 33.31\small{\textpm0.33} & 9.74\small{\textpm0.50} & 22.61 \\

    w/o. Mention Classifier & 13.68\small{\textpm0.35} & 9.13\small{\textpm0.30} & 11.10\small{\textpm0.35} & 8.21\small{\textpm0.23} & 10.53 & 6.17\small{\textpm0.29} & 5.02\small{\textpm0.31} & 5.81\small{\textpm0.31} & 7.32\small{\textpm0.20} & 6.08 \\

    w/o. Margin-based Learning & 32.11\small{\textpm0.16} & 29.10\small{\textpm0.20} & 30.32\small{\textpm0.31} & 31.99\small{\textpm0.13} & 30.88 & 40.07\small{\textpm0.20} & 43.50\small{\textpm0.29} & 44.80\small{\textpm0.31} & 45.11\small{\textpm0.27} & 43.37 \\
    \bottomrule
    \end{tabular}
    }
    \caption{F1 scores with standard deviations on CrossNER. $^{\ddag}$ denotes the results reported in~\citet{Hou2020Few}.}
    \label{tab:ablation-2}
\end{table*}

\section{Details of Our Variants}
\label{app:variants}

\noindent\textbf{{\model} w/o. Span Extractor.} We remove the span extractor. We leverage the standard prototypical network to perform the token-level type classification, which is the same as ProtoBERT.

\begin{table}
\centering
\resizebox{\linewidth}{!}{
\begin{tabular}{lc}
\toprule
\bf Hyper-parameter &\bf Value \\
\midrule
Batch Size & \{1, 2, 4, 8\} \\
Learning Rate & \{1e-5, 2e-5, 5e-5, 1e-4, 2e-4\} \\
Dropout Rate & \{0.1, 0.3, 0.5\} \\
$T'$ & \{0, 100, 200, 500, 1,000\} \\
$\theta$ & \{0.0, 0.50, 0.75, 0.80, 0.85, 0.90, 0.95\} \\
$r$ & \{1, 2, 3, 4, 5, 6\} \\
\bottomrule
\end{tabular}
}
\caption{The searching scope for each hyper-parameter.}
\label{tab:search-scope}
\end{table}

\noindent\textbf{{\model} w/o. Mention Classifier.} 
We remove the mention classifier. To support the classification, we directly use the K-Means algorithm. 
Specifically, we first train the span extractor on the support set and extract all candidates for each query example. We then obtain the span embedding based on the contextual representations. Thus, for each episode data, we can obtain all span representations, we leverage the K-Means to obtain $N$ cluster on all support spans, and then predict each type of the query span based on the nearest cluster.

\noindent\textbf{{\model} w/o. Margin-based Learning.} 
We remove the margin-based learning objective to validate its contribution. Therefore, we do not obtain the false positives from the query set in the training episode. 
Specifically, we first obtain all candidates for each query example, and then directly classify each span based on the prototypical network.

\section{Details of the Grid Search}
\label{app:grid-search}
The searching scope of each hyper-parameter is shown in Table~\ref{tab:search-scope}. Note that, the batch size in the $N$-way $K$-shot setting means the number of episode data in one batch.

\section{Details of the Ablation Study}
\label{app:ablation}
The detail results of ablation study are listed in Table~\ref{tab:ablation-1} and Table~\ref{tab:ablation-2}.

\end{document}